\documentclass{article}

\usepackage{arxiv}
\usepackage[utf8]{inputenc}
\usepackage[T1]{fontenc}
\usepackage{url}
\usepackage{booktabs}
\usepackage{amsfonts}
\usepackage{nicefrac}
\usepackage{microtype}
\usepackage{graphicx}
\usepackage{natbib}
\usepackage{doi}
\usepackage{xcolor}
\usepackage[table]{xcolor}
\usepackage{subcaption}
\usepackage{amsmath}
\usepackage{multirow}
\usepackage{authblk}
\usepackage{hyperref}
\usepackage{authblk}

\title{Coronary Artery Segmentation and Vessel-Type Classification in X-Ray Angiography}

\newcommand*\samethanks[1][\value{footnote}]{\footnotemark[#1]}

\author[1,2]{\href{mailto:yousefzadeh.meh@gmail.com}{Mehdi Yousefzadeh}}
\author[3]{Siavash Shirzadeh Barough}
\author[4]{Ashkan Fakharifar}
\author[2]{Yashar Tayyarazad}
\author[2]{Narges Eghbali}
\author[2]{Mohaddeseh Mozaffari}
\author[5]{Hoda Taeb}
\author[6]{Negar Sadat Rafiee Tabatabaee}
\author[1]{\href{mailto:parsa.esfahanian@ipm.ir}{Parsa Esfahanian}}
\author[7]{Ghazaleh Sadeghi Gohar}
\author[7]{Amineh Safavirad}
\author[7]{Saeideh Mazloomzadeh}
\author[7]{\href{mailto:Ehsankhalilipur@gmail.com}{Ehsan khalilipur}}
\author[7]{\href{mailto:Arelahifar@gmail.com}{Armin Elahifar}\thanks{Corresponding authors}\,\,}
\author[7]{{Majid Maleki}\samethanks\,\,}

\affil[1]{School of Computer Science, Institute for Research in Fundamental Sciences (IPM), Tehran, Iran}
\affil[2]{Department of Physics, Shahid Beheshti University, Tehran, Iran}
\affil[3]{Brain Mapping Research Center, Shahid Beheshti University of Medical Sciences, Tehran, Iran}
\affil[4]{Gilan University of Medical Sciences, Rasht, Gilan, Iran}
\affil[5]{Department of Physics, Simon Fraser University, Burnaby, Canada}
\affil[6]{Fatemeh Zahra Hospital, Alborz University of Medical Sciences, Eshtehard, Iran}
\affil[7]{Rajaie Cardiovascular Medical and Research Institute, Iran University of Medical Sciences, Tehran, Iran}

\renewcommand{\shorttitle}{Coronary Artery Segmentation and ...}

\hypersetup{
    pdftitle={Coronary Artery Segmentation and Vessel-Type Classification in X-Ray Angiography},
    pdfsubject={cs.CV, eess.IV},
    pdfauthor={Mehdi Yousefzadeh, Siavash Shirzadeh Barough, Ashkan Fakharifar, Yashar Tayyarazad, Narges Eghbali, Mohaddeseh Mozaffari, Hoda Taeb, Negar Sadat Rafiee Tabatabaee, Parsa Esfahanian, Ghazaleh Sadeghi Gohar, Amineh Safavirad, Armin Elahifar, Saeideh Mazloomzadeh, Ehsan Khalilipur, Majid Maleki},
    pdfkeywords={X-Ray Angiography, Coronary Arteries Segmentation, Vessel-Type Labeling, Fractional Flow Reserve},
}

\begin{document}
\maketitle
\setcounter{footnote}{0}

\begin{abstract}
\textbf{Introduction.}
X-ray coronary angiography (XCA) is the clinical reference standard for assessing coronary artery disease, yet quantitative analysis is limited by the difficulty of robust vessel segmentation in routine data. Low contrast, motion, foreshortening, overlap, and catheter confounding degrade segmentation and contribute to domain shift across centers. Reliable segmentation, together with vessel-type labeling, enables vessel-specific coronary analytics and downstream measurements that depend on anatomical localization.

\textbf{Methods.}
From 670 cine sequences (407 subjects), we select a best frame near peak opacification using a low-intensity histogram criterion and apply joint super-resolution and enhancement. We benchmark classical Meijering/Frangi/Sato pipelines under per-image oracle tuning, a single global mean setting, and per-image prediction via Support Vector Regression (SVR) from a 140-D descriptor. Neural baselines include U-Net and FPN (SE-ResNet18/SE-ResNeXt50) and a Swin-Transformer at $384\times384$ or $756\times756$, trained with coronary-only, catheter-only, and merged coronary+catheter supervision. A second stage assigns vessel identity (LAD/LCX/RCA). External evaluation uses the public DCA1 cohort.

\textbf{Results.}
SVR per-image tuning improves Dice over global means for all classical filters (Meijering: 0.703 vs.\ 0.698; Sato: 0.719 vs.\ 0.710; Frangi: 0.759 vs.\ 0.741). Among deep models, FPN+SE-ResNet18 at $756\times756$ attains $0.914 \pm 0.007$ (coronary-only), and merged coronary+catheter labels further improve to $0.931 \pm 0.006$ (catheter-only: $0.856 \pm 0.031$). On DCA1 as a strict external test, Dice drops to 0.798 (coronary-only) and 0.814 (merged), while light in-domain fine-tuning recovers to $0.881 \pm 0.014$ and $0.882 \pm 0.015$. Vessel-type labeling achieves accuracy/DSC of 98.5\%/0.844 (RCA), 95.4\%/0.786 (LAD), and 96.2\%/0.794 (LCX).

\textbf{Conclusion.}
Machine-learning generalized image processing and deep neural segmentation provide accurate coronary segmentation and vessel-type labeling in XCA. Learned per-image tuning strengthens classical pipelines, while high-resolution FPN models and merged-label supervision improve stability and external transfer with modest adaptation.
\end{abstract}

\keywords{X-Ray Angiography \and Coronary Arteries Segmentation \and Vessel-Type Labeling \and Fractional Flow Reserve}

\section{Introduction}
\label{sec:introduction}

Coronary angiography is the clinical reference standard for diagnosing and characterizing coronary artery disease (CAD), providing high-fidelity visualization of lumen narrowing to guide revascularization and other life-saving interventions \cite{fihn20142014}. It plays a central role across a wide range of CAD presentations, including acute myocardial infarction, unstable angina, and ischemic heart disease, and it informs the choice between percutaneous coronary intervention (PCI) and coronary artery bypass grafting (CABG) \cite{patel2017acc}.

Cardiovascular disease remains a leading global cause of death, and the burden of CAD motivates robust, scalable tools for angiographic assessment \cite{tsao2023heart}.

Coronary artery segmentation is a core component of quantitative angiographic analysis. It enables objective assessment of stenosis severity, supports visualization of contrast flow patterns, and provides a foundation for downstream functional estimates such as fractional flow reserve (FFR) surrogates \cite{morris2013virtual,xu2017diagnostic}.

In practice, accurate segmentation is challenging because angiograms often exhibit low contrast, motion artifacts, foreshortening, and vessel overlap. Interventional devices such as catheters can further confound vessel delineation, increasing the difficulty of both manual assessment and automated methods \cite{reiber1985assessment,zir1976interobserver}. These challenges contribute to variability among readers and can be particularly consequential in borderline lesions.

A range of classical approaches has been explored to improve vessel visibility prior to segmentation. Statistical methods have been used to describe intensity distributions through histogram-based descriptors and probability density estimation, with the goal of enhancing contrast and outlining vessel boundaries \cite{Roy2023,Hwang2020}. Other work has quantified spatial dependencies in vascular trees using correlation functions across scales \cite{Sabuncu2006,HOG2020}.

Entropy-based criteria, including maximum-entropy approaches and generalized divergences such as Jensen--Shannon, have been applied to capture image complexity and guide adaptive thresholding \cite{MEM2021,Kasai2023}. These descriptors often appear in multi-stage pipelines that combine enhancement operations such as unsharp masking and CLAHE with filter-based feature extraction \cite{Preproc2021,GBDT2021}. More recently, non-parametric density estimation and adaptive entropy imaging have been proposed to mitigate histogram binning limitations and improve delineation of small vessels \cite{CoronaryDominance2025,Ultrasound2023}.

Deep learning has substantially advanced automated segmentation in medical imaging through end-to-end feature learning and multi-scale representations \cite{Litjens2017}. U-Net and its variants introduced encoder--decoder designs with skip connections that preserve spatial detail under limited annotation budgets \cite{Ronneberger2015,Zhou2018}. Auto-configuring pipelines such as nnU-Net demonstrated strong performance across many segmentation tasks by adapting training and architectural choices to the dataset \cite{Isensee2021}.

Attention mechanisms and hybrid CNN--Transformer models further improved performance by emphasizing salient anatomy and capturing longer-range context \cite{Oktay2018,Cai2023,Liu2021Swin}. In coronary angiography, these models can be highly accurate, but performance can depend strongly on input resolution, supervision design, and domain shift across centers and acquisition protocols.

Despite progress, coronary angiography segmentation remains difficult to standardize. Many published results rely on small cohorts, different preprocessing choices, and inconsistent evaluation protocols. Public datasets such as DCA1 provide valuable benchmarks, but they typically offer binary vessel masks, which limits analysis of device interference and does not support vessel-level labeling required for vessel-specific downstream analysis \cite{DCA1}.

In parallel, classical vesselness filters based on the Hessian matrix remain relevant due to their strong inductive bias for tubular structures. Methods such as Frangi, Meijering, and Sato have been widely used, and broader benchmarking across modalities has shown that their performance is sensitive to parameter choices and imaging characteristics \cite{frangi1998multiscale,meijering2004neurite,sato19973d,lamy2022benchmark}. In coronary angiography specifically, Hessian-based pipelines have been used in combination with enhancement, region merging, and region growth strategies, reporting competitive Dice values in some settings \cite{wan2018automated,kerkeni2016coronary,ma2020coronary}.

This work aims to provide a unified and practical study of coronary segmentation and vessel-type classification in X-ray coronary angiography (XCA). We curate an expert-reviewed dataset with separate coronary and catheter masks, as well as vessel-level labels for the major coronaries (LAD, LCX, RCA).

We benchmark classical vesselness pipelines and modern deep models under a consistent preprocessing and evaluation protocol. For classical methods, we go beyond fixed global hyperparameters by learning to predict per-image filter settings from inexpensive image descriptors using support vector regression (SVR), which improves robustness without requiring ground-truth masks at inference time.

For deep models, we compare U-Net, feature pyramid networks (FPN), and a hierarchical Transformer baseline, and we study the impact of supervision design by training with coronary-only, catheter-only, and merged coronary+catheter targets. Finally, we introduce a second-stage model that assigns vessel identity within the segmented coronary tree, bridging segmentation outputs with vessel-specific downstream analysis.

Here are the key contributions of our work:
\begin{itemize}
    \item Curated a new dataset with expert-reviewed labels, including separate masks for coronary arteries and catheter, plus vessel-level masks (LAD, LCX, RCA).
    \item Systematically benchmarked classical vesselness pipelines against each other and against neural segmentation models under a unified protocol.
    \item Enhanced classical pipelines via a learning-to-tune strategy that predicts per-image hyperparameters using SVR, improving robustness without ground-truth labels at inference.
    \item Conducted a controlled study of label design, comparing coronary-only, catheter-only, and merged coronary+catheter supervision, and quantified the impact of device interference.
    \item Achieved strong Dice similarity on coronary segmentation with high-resolution multi-scale models, and characterized domain shift using external evaluation on DCA1 \cite{DCA1}.
    \item Introduced a second-stage vessel-type labeling model for LAD, LCX, and RCA to enable vessel-specific downstream analyses.
\end{itemize}

The remainder of this paper is organized as follows. Section~\ref{sec:method} describes dataset curation, best-frame selection, annotation workflow, and preprocessing, followed by classical learning-to-tune pipelines and deep segmentation models, including vessel-type labeling. Section~\ref{sec:results} presents internal and external evaluations, comparisons, and ablations on label design and resolution. Section~\ref{sec:dis} discusses error modes, clinical interpretation, limitations, and future directions.

\section{Method and Data}
\label{sec:method}

\begin{figure}[t]
  \centering
  \includegraphics[width=\textwidth]{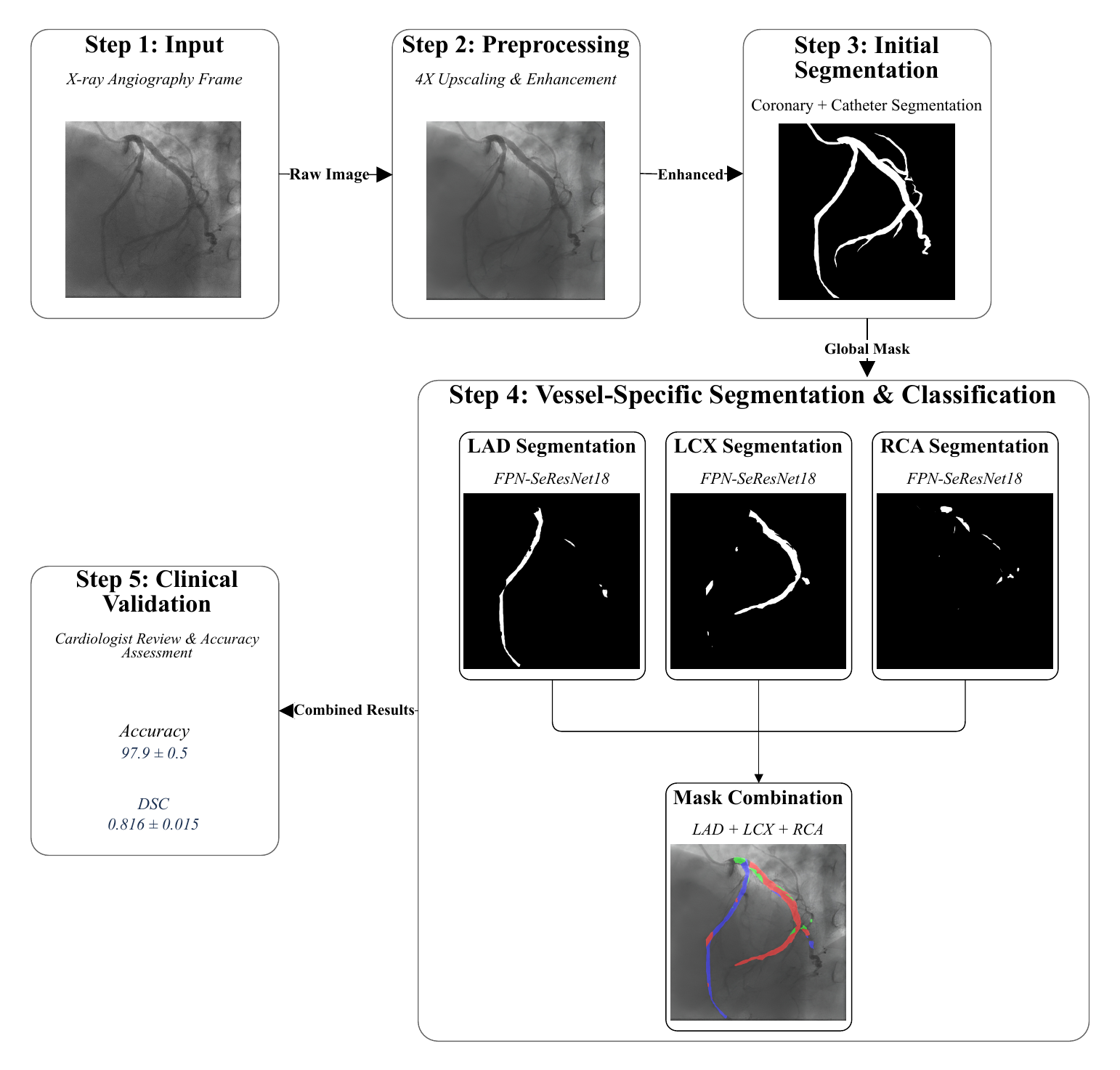}
  \caption{Overview of the proposed XCA analysis workflow. From each cine, a best frame is selected near peak opacification and enhanced, followed by first-stage segmentation to obtain a global foreground mask. A second stage assigns vessel identity within the coronary tree, and outputs are reviewed by an interventional cardiologist for clinical consistency.}
  \label{fig:framework}
\end{figure}

In this section, we describe the data sources, annotation workflow, and preprocessing pipeline used throughout our experiments, followed by the segmentation and classification methods evaluated in this work. Figure~\ref{fig:framework} summarizes the overall workflow, from best-frame selection and image enhancement to first-stage coronary segmentation and second-stage vessel-type labeling.

We first introduce our curated clinical cohort and labeling protocol, and then describe the public DCA1 dataset used for external evaluation. We then present two complementary segmentation families: (i) classical vesselness filters with learned per-image hyperparameter prediction, and (ii) deep neural models (U-Net, FPN, and Swin-Transformer) trained under different label regimes, followed by a dedicated model for LAD/LCX/RCA assignment within the predicted coronary tree.

\subsection{Dataset Description}
\label{sec:dataset}

\subsubsection{Cohort and dataset summary}
In total, 670 XCA cine sequences were gathered from 407 unique subjects. Clinical decision-making information was available for a subset of the cohort. Per the clinical records, percutaneous coronary intervention (PCI) was recommended for 105 out of 390 patients with available outcome annotations.

The cohort reflects routine acquisition variability, including differences in contrast opacification, background clutter, and vessel overlap across views. This diversity is important for assessing segmentation robustness under realistic conditions.

\begin{figure}[b]
  \centering
  \includegraphics[width=\textwidth]{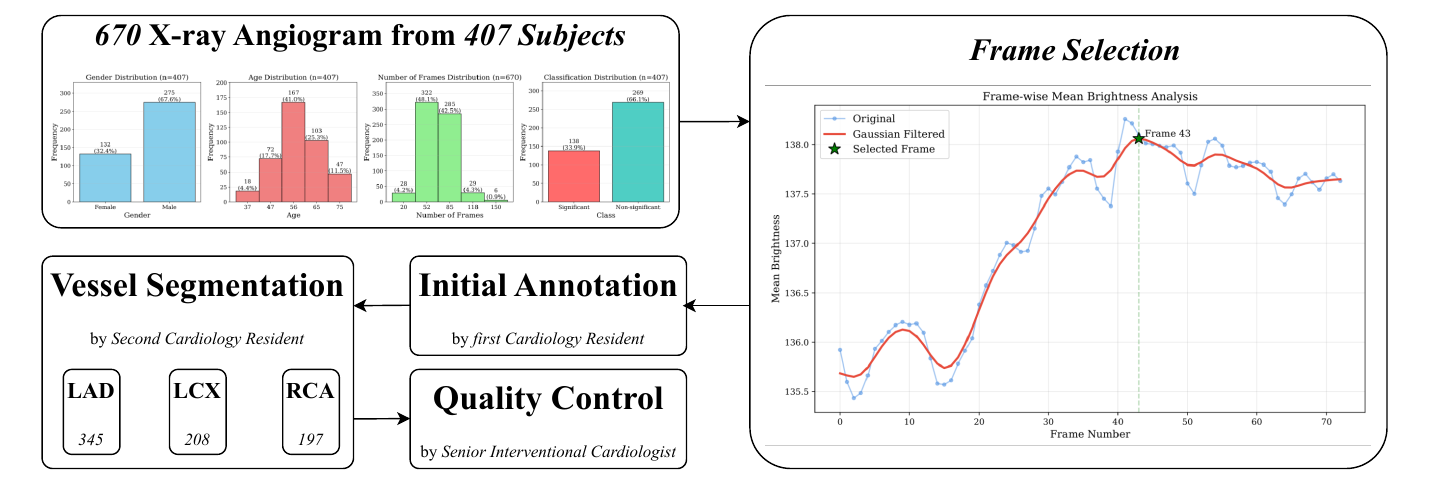}
  \caption{Dataset and labeling summary. Best frames are extracted from cine angiograms and annotated to produce coronary-artery and catheter masks, with additional vessel-level labels for the major coronaries when visible. All masks were reviewed and corrected as needed by a senior interventional cardiologist prior to final approval.}
  \label{fig:datas}
\end{figure}

To minimize procedure-related artifacts, all sequences used in this study were selected from clinically good views and recorded prior to any therapeutic manipulation. Descriptive characteristics of the cohort (sex, age distribution) and the empirical distribution of the number of frames per angiography are summarized in Figure~\ref{fig:datas}.

All data were collected under institutional ethics approval (protocol code IR.RHC.REC.1403.038) at the Shahid Rajaie Cardiovascular Research and Treatment Institute, Tehran, Iran. XCA was performed using a Philips angiography system. Each cine sequence was acquired at a spatial resolution of $512 \times 512$ pixels with a frame rate of 15 frames per second (FPS).

\subsubsection{Best-frame selection from cine angiography}
\label{subsec:bestframe}
Our objective was to extract, from each cine sequence, a single 2D frame captured near systole in which the coronary tree presents the largest visible extent. In practice, this typically coincides with the time point at which the contrast medium produces the highest degree of vascular opacification, when a larger fraction of pixels take on low grayscale intensities (darker appearance).

Because opacification evolves over time within each cine, a randomly chosen frame may underrepresent distal vessels or yield ambiguous boundaries. We therefore use a simple, deterministic best-frame heuristic that can be applied consistently across the cohort.

Let $I_t \in [0,1]^{512\times512}$ denote the intensity-normalized frame at time index $t$ and let $h_t(k)$ be the empirical histogram (or probability mass function) of pixel intensities for $I_t$ over $K$ bins. We define a low-intensity range $\mathcal{L}$ (corresponding to darker pixels dominated by contrast) and use the following criterion:
\begin{equation}
t^\star \;=\; \operatorname*{arg\,max}_{t} \; \max_{k \in \mathcal{L}} \; h_t(k),
\end{equation}
that is, we select the frame whose intensity histogram exhibits the highest peak within the low-intensity region. In implementation, $\mathcal{L}$ can be set as the lower quantile range of the per-frame intensity distribution. We used a fixed low-intensity band across frames for robustness.

This heuristic tracks the per-frame intensity distribution throughout the cine and selects the frame with the most pronounced mode at low intensities. Empirically, this matches the desired phase with maximal contrast opacification and vessel visibility.

\newpage

\subsubsection{Annotation workflow}
Figure~\ref{fig:datas} depicts the end-to-end pipeline for data selection and labeling. The labeling process proceeded in four stages.

\paragraph{Stage 1: Initial semantic segmentation.}
From each cine, the best frame (Section~\ref{subsec:bestframe}) was extracted, yielding 670 frames for manual annotation. A cardiology resident produced pixel-wise segmentation masks with two primary classes: (i) coronary arteries (including main vessels and visible side branches) and (ii) catheter. Annotations were made on each frame independently.

\paragraph{Stage 2: Major-vessel landmarking and review.}
A second cardiology resident independently reviewed the masks and annotated proximal and distal landmarks (start and end points) of the three major coronaries: left anterior descending (LAD), left circumflex (LCX), and right coronary artery (RCA). In total, 345 LAD, 208 LCX, and 197 RCA masks were identified. In some frames, both LAD and LCX masks were present (depending on view and opacification).

\paragraph{Stage 3: Vessel-specific mask construction.}
Using the expert-provided landmarks, the technical team constructed three vessel-specific masks by filling the regions between corresponding start and end landmarks, resulting in distinct masks for LAD, LCX, and RCA. These were retained in addition to the global coronary-artery mask and the catheter mask.

\paragraph{Stage 4: Senior interventional review and ground-truth consolidation.}
All masks (global and vessel-specific) underwent final adjudication by a senior interventional cardiologist, who applied corrections where necessary and provided final approval. The resulting ground truth comprises (a) primary classes \texttt{catheter} and \texttt{coronary\_arteries}, and (b) a vessel-level decomposition of \texttt{coronary\_arteries} into \texttt{LAD}, \texttt{LCX}, \texttt{RCA}, and remaining coronary regions when visible.

\subsubsection{Public external dataset (Mexico; DCA1)}
For external validation and testing, we used the publicly released \emph{Database of X-ray Coronary Angiograms} (DCA1), curated by the Cardiology Department of the Mexican Social Security Institute (IMSS, UMAE T1--León) and hosted by CIMAT \cite{DCA1}. DCA1 consists of expert-annotated X-ray coronary angiograms paired with binary vessel ground-truth masks. Images are distributed as grayscale PGM files with a native size of approximately $300\times300$ pixels.

The original release describes a benchmark split of 100 training and 30 test images and is widely used in the literature for vessel-segmentation benchmarking \cite{CervantesSanchez2019AppliedSci,Tao2022Frontiers}. In our experiments (see Results), given that catheters in DCA1 are often faint or partially cropped and due to differences in labeling standards, we evaluate two settings on this dataset (coronary-only and merged coronary+catheter).

\subsubsection{Preprocessing and image enhancement}
\label{subsec:preprocessing}
Raw frames were acquired at $512 \times 512$ pixels. Because some segmentation backbones benefit from higher-resolution inputs and reduced noise, we applied a convolutional neural network that performs joint $4\times$ super-resolution and image enhancement (denoising and contrast preservation) in a single pass.

For downstream experiments, we generated two analysis scales: $384 \times 384$ and $756 \times 756$. This provided compatibility with architectures expecting smaller receptive fields as well as those favoring higher spatial detail, balancing memory and throughput constraints.

For all models, images were converted to single-channel grayscale if not already in that format, and pixel intensities were linearly normalized to the $[0,1]$ range. Unless otherwise stated, all training and evaluation used these normalized tensors as network inputs.

\subsection{Segmentation methods}

\subsubsection{Vessel enhancement filters}
\label{sec:vessel_filters}
We use three filters, Meijering, Frangi, and Sato, to enhance blood vessels in our images. These filters operate on the Hessian matrix, which captures second-order structure. The filters are applied at multiple scales by convolving the image with Gaussian kernels of varying standard deviations $\sigma$.

For a 2D image, the Hessian matrix is given by:
\begin{equation}
H_{ij}(x, \sigma) = \frac{\partial^2}{\partial i \partial j} \big[I(x) * G(x, \sigma)\big], \quad i,j\in\{x,y\},
\end{equation}
where
\(
G(x, \sigma) = \frac{1}{2\pi\sigma^2}\exp\left(-\frac{x^2+y^2}{2\sigma^2}\right)
\)
is the Gaussian kernel, $I(x)$ represents the image intensity, and $*$ indicates convolution \cite{frangi1998multiscale}. The eigenvalues of the Hessian matrix, $\lambda_1$ and $\lambda_2$, are used to distinguish vessel-like regions from background. The three filters differ in their vesselness response function $V(x,\sigma)$ and how they combine the Hessian eigenvalues \cite{sato19973d}.

\paragraph{Meijering filter.}
The Meijering filter uses a linear combination of eigenvalues. For a 2D image, it evaluates $\lambda_1$ and $\lambda_2$, where $|\lambda_1| > |\lambda_2|$, and defines:
\begin{equation}
V(x, \sigma) = \max\left(0, \lambda_1 + \alpha \lambda_2\right),
\end{equation}
where $\alpha$ is a shaping parameter, typically set to $-\frac{1}{2}$ for 2D images \cite{meijering2004neurite}. This formulation is computationally efficient, but it may be more sensitive to noise because it does not explicitly suppress non-vessel structures.

\paragraph{Frangi filter.}
The Frangi filter enhances tubular structures while suppressing non-tubular regions through exponential terms. It evaluates $\lambda_1$ and $\lambda_2$ (with $|\lambda_1| > |\lambda_2|$) and defines:
\begin{equation}
\begin{aligned}
V(x, \sigma) &=
\begin{cases}
0 & \text{if } \lambda_2 > 0, \\
\left(1 - e^{-\frac{R_A^2}{2\alpha^2}}\right)\left(1 - e^{-\frac{S^2}{2\beta^2}}\right) & \text{otherwise},
\end{cases}
\end{aligned}
\end{equation}
where $R_A = \frac{|\lambda_2|}{|\lambda_1|}$ measures elongation and $S = \sqrt{\lambda_1^2 + \lambda_2^2}$ measures overall structure strength. The parameters $\alpha$ and $\beta$ control sensitivity \cite{frangi1998multiscale}.

\paragraph{Sato filter.}
The Sato filter follows a related approach and introduces additional shape descriptors. It evaluates $\lambda_1$ and $\lambda_2$ (with $|\lambda_1| < |\lambda_2|$) and defines:
\begin{equation}
\begin{aligned}
V(x, \sigma) &=
\begin{cases}
0 & \text{if } \lambda_2 \leq 0, \\
\left(1 - e^{-\frac{R_A^2}{2\sigma^2}}\right) e^{-\frac{R_B^2}{2\sigma^2}} \left(1 - e^{-\frac{S^2}{2\sigma^2}}\right) & \text{otherwise},
\end{cases}
\end{aligned}
\end{equation}
where $R_A = \frac{|\lambda_1|}{|\lambda_2|}$, $R_B = \frac{|\lambda_1 + \lambda_2|}{\sqrt{\lambda_1^2 + \lambda_2^2}}$, and $S = \sqrt{\lambda_1^2 + \lambda_2^2}$ \cite{sato19973d}.

\paragraph{Preprocessing and postprocessing.}
Before applying vesselness filters, we use the preprocessing and enhancement described in Section~\ref{subsec:preprocessing}. Angiographic frames often contain dark borders that can be misidentified as vessels. To reduce these false positives, we examine rows and columns within a 100-pixel boundary band and compute the mean pixel intensity for each candidate row/column on the native 8-bit grayscale scale (0--255). Rows or columns with mean intensity below 50 are removed, and the procedure is applied symmetrically on all sides to eliminate black margins while preserving vessel structures.

Postprocessing binarizes the filter response using a tuned threshold to maximize Dice. Two common errors are then addressed: (i) small gaps within vessels are filled using \texttt{binary\_closing} \cite{van2014scikit}, with the structuring-element size chosen to avoid spurious connections between nearby branches, and (ii) small false-positive regions are removed using \texttt{remove\_small\_objects} \cite{van2014scikit}, with the minimum component size tuned to suppress background noise.

\paragraph{Hyperparameter tuning.}
To apply the filters, we performed a grid search to tune hyperparameters and maximize Dice. The search ranges for each filter are listed in Table~\ref{tab:hyperparameters}.

\begin{table}[t]
    \centering
    \caption{Grid-search ranges for hyperparameters of the Meijering, Sato, and Frangi vesselness pipelines, including binarization and postprocessing settings.}
    \begin{tabular}{|c|c|c|}
        \hline
        \textbf{Filter} & \textbf{Hyperparameter} & \textbf{Optimal Range} \\ 
        \hline
        \multirow{4}{*}{Meijering}
            &\cellcolor[gray]{0.8} $\sigma$ (Sigma) & \cellcolor[gray]{0.8}2.5 to 5.5 (step size: 0.5) \\ \cline{2-3}
            & Binarization Threshold & 0.05 to 0.12 (step size: 0.01) \\ \cline{2-3}
            & \cellcolor[gray]{0.8}Disk Size (Binary Closing) & \cellcolor[gray]{0.8}1, 2, 3 \\ \cline{2-3}
            & Minimum Region Size & 100 to 4000 (step size: 50) \\ \cline{2-3}
        \hline
        \multirow{4}{*}{Sato}
            & \cellcolor[gray]{0.8}$\sigma$ (Sigma) & \cellcolor[gray]{0.8}2.0 to 6.5 (step size: 0.5) \\ \cline{2-3}
            & Binarization Threshold & 0.010 to 0.031 (step size: 0.001) \\ \cline{2-3}
            & \cellcolor[gray]{0.8}Disk Size (Binary Closing) & \cellcolor[gray]{0.8}1, 2, 3, 4, 5 \\ \cline{2-3}
            & Minimum Region Size & 100 to 5500 (step size: 50) \\ \cline{2-3}
        \hline
        \multirow{6}{*}{Frangi}
            & \cellcolor[gray]{0.8}$\sigma$ (Sigma) & \cellcolor[gray]{0.8}1.0 to 6.0 (step size: 0.5) \\ \cline{2-3}
            & Binarization Threshold & 0.3 to 0.7 (step size: 0.05)\\ \cline{2-3}
            & \cellcolor[gray]{0.8}$\alpha$ (Alpha) & \cellcolor[gray]{0.8}0.5 to 1.0 (step size: 0.05)\\ \cline{2-3}
            & $\beta$ (Beta) & 0.5 to 1.0 (step size: 0.05) \\ \cline{2-3}
            & \cellcolor[gray]{0.8}Maximum Hole Area & \cellcolor[gray]{0.8}200 to 500 (step size: 10) \\ \cline{2-3}
            & Minimum vessel Area & 50 to 100(step size: 5) \\ \cline{2-3}
        \hline
    \end{tabular}
    \label{tab:hyperparameters}
\end{table}

\subsubsection{Applying machine learning to per-image parameter prediction}
Classical vesselness pipelines (Section~\ref{sec:vessel_filters}) require hyperparameters (for example, scale $\sigma$, response thresholds, and morphology settings) that we initially selected per image by an oracle grid search to maximize Dice against the ground-truth mask. Concretely, for filter $m \in \{\text{Meijering},\text{Frangi},\text{Sato}\}$ with parameter vector $\boldsymbol{\theta}_m$, the oracle chooses
\begin{equation}
\boldsymbol{\theta}_m^{\star}(I)\;=\;\operatorname*{arg\,max}_{\boldsymbol{\theta}_m\in\Theta_m}\;
\mathrm{DSC}\!\big(\,F_m(I;\boldsymbol{\theta}_m),\,M_{\mathrm{gt}}\,\big),
\end{equation}
where $F_m$ denotes the full filter-to-threshold-to-morphology pipeline (Section~\ref{sec:vessel_filters}). At test time, $M_{\mathrm{gt}}$ is unavailable, so using a single global setting (or the mean of per-image optima) degrades accuracy.

Instead, we learn a regression from inexpensive image descriptors to oracle parameters, enabling per-image adaptation without access to ground truth.

\paragraph{Feature design (140-D) and learning target.}
For each angiography frame $I$, we compute a fixed-length descriptor $\boldsymbol{\phi}(I)\in\mathbb{R}^{140}$ by concatenating three families of statistics:
(i) an intensity histogram $p(k)$ over gray levels;
(ii) Minkowski functionals of superlevel sets, which summarize morphology. For a threshold $\tau$, form $B_\tau=\{\mathbf{x}\mid I(\mathbf{x})\ge\tau\}$ and measure area $A(\tau)$, boundary length $L(\tau)$, and Euler characteristic $\chi(\tau)$ \cite{Mecke2000,Michielsen2001};
(iii) topological data analysis (TDA) summaries via Betti curves $\beta_0(\tau)$, $\beta_1(\tau)$, and $\beta_2(\tau)$ computed on the cubical-complex filtration of $I$ \cite{Carlsson2009,Edelsbrunner2010}.

Each curve is sampled on a uniform grid and concatenated (with min-max normalization) to produce a 140-D vector. We use 8-connectivity for $\chi$ and $\beta_0$/$\beta_1$ on binary superlevel sets. Results were stable under 4-connectivity as well.

\paragraph{Support Vector Regression (SVR) for hyperparameter prediction.}
We train independent $\varepsilon$-SVR models with RBF kernels to map $\boldsymbol{\phi}(I)$ to each scalar entry of $\boldsymbol{\theta}_m^{\star}(I)$, yielding one regressor per hyperparameter per filter \cite{Smola2004,Drucker1997}. Targets are standardized (and log-transformed for strictly positive scales), features are z-scored, and $(C,\varepsilon,\gamma)$ are selected by nested cross-validation on the training set of images endowed with oracle labels $\boldsymbol{\theta}_m^{\star}$.

At inference, for a new image $I'$ we compute $\boldsymbol{\phi}(I')$, predict $\widehat{\boldsymbol{\theta}}_m$, clip to the admissible bounds $\Theta_m$, and snap to the nearest grid step used during oracle search (Table~\ref{tab:hyperparameters}). The predicted settings are then fed into the corresponding vesselness pipeline (filtering, binarization, hole-filling, and small-object removal) to produce a segmentation mask.

\subsubsection{CNN- and Transformer-based segmentation architectures}
We employed six segmentation configurations spanning encoder--decoder CNNs (U-Net and FPN) and a hierarchical Vision Transformer (Swin). All convolutional backbones were initialized from ImageNet pretraining \cite{Deng2009} and augmented with Squeeze-and-Excitation (SE) blocks to adaptively reweight channels. Inputs were grayscale, intensity-normalized images at either $384\times384$ or $756\times756$, derived from the preprocessing pipeline described earlier.

\paragraph{U-Net with SE-ResNet18.}
A standard U-Net topology with an SE-ResNet18 encoder \cite{He2016} was used, preserving long skip connections from encoder stages to the symmetric decoder to recover fine vessel detail. SE modules were inserted after each residual block to emphasize low-contrast vascular signal and suppress background clutter \cite{Ronneberger2015,Hu2018}.

\paragraph{U-Net with SE-ResNeXt50.}
We replaced the encoder with SE-ResNeXt50 to leverage grouped convolutions (higher cardinality) for richer features while retaining the same U-Net decoder and skip connections. SE blocks were interleaved within each ResNeXt stage for dynamic channel recalibration \cite{Xie2017,Hu2018,Ronneberger2015}.

\newpage

\paragraph{FPN with SE-ResNet18.}
Feature Pyramid Networks aggregate lateral encoder features via a top-down path to form multi-scale pyramids. Using an SE-ResNet18 encoder \cite{He2016}, lateral $1\times1$ projections and top-down upsampling produced semantically strong, high-resolution maps for the final segmentation head \cite{Lin2017,Hu2018}.

\paragraph{FPN with SE-ResNeXt50.}
This variant keeps the same FPN design but uses an SE-ResNeXt50 encoder, increasing representational capacity through grouped convolutions while maintaining the same multi-scale fusion strategy \cite{Xie2017,Lin2017,Hu2018}.

\paragraph{Swin-Transformer.}
As a non-convolutional baseline, we used a hierarchical Swin-Transformer encoder with shifted-window self-attention and a lightweight upsampling head to produce dense masks. The hierarchical design preserves locality while capturing longer-range dependencies relevant for tortuous coronary anatomy \cite{Liu2021Swin}.

\paragraph{Label regimes and inputs.}
Each architecture was trained under three supervision regimes using the same network definitions and hyperparameters: (i) a coronary-only mask (all visible coronary arteries as foreground), (ii) a catheter-only mask, and (iii) a merged coronary+catheter mask where both structures form a single foreground. In all cases, the input is the preprocessed angiography frame and the training target is a binary mask corresponding to the chosen regime.

\paragraph{Second-stage vessel labeling (LAD/LCX/RCA).}
Downstream, we derive vessel-specific labels from the output mask of the best-performing first-stage model. Concretely, three vessel-specific binary segmentation models are trained to predict per-pixel posteriors for the main coronaries: $p_\text{LAD}(\mathbf{x})$, $p_\text{LCX}(\mathbf{x})$, and $p_\text{RCA}(\mathbf{x})$.

Their inputs are the first-stage foreground mask (optionally concatenated with the grayscale image), enabling the second stage to specialize in topological and positional cues without relearning vessel-background separation. At inference, we gate predictions by the first-stage mask and assign each foreground pixel the class with the highest probability,
\begin{equation}
\hat{c}(\mathbf{x}) \;=\; \operatorname*{arg\,max}_{c \in \{\mathrm{LAD},\,\mathrm{LCX},\,\mathrm{RCA}\}} \; p_c(\mathbf{x}),
\end{equation}
followed by small-component removal and contour smoothing to encourage anatomically plausible outputs.

\subsection{Training and evaluation protocol}
All experiments used five-fold cross-validation to estimate performance. In each fold, data were split into training and validation sets with no subject-level overlap. We report results as the mean and standard deviation across folds.

Deep models were trained for up to 100 epochs using the Adam optimizer and a Dice-based loss, with a batch size of 16. We reduced the learning rate when validation performance plateaued and applied early stopping based on validation Dice to select the final checkpoint. All models used the same preprocessing pipeline described in Section~\ref{subsec:preprocessing}.

External evaluation was performed on DCA1 under two settings: (i) zero-shot testing (no adaptation) and (ii) light in-domain fine-tuning using the DCA1 training split, with results reported on the held-out DCA1 test split. Fine-tuning used the same loss and optimizer as internal training and was run for a small number of epochs with early stopping on a validation subset.

\section{Results}
\label{sec:results}

\subsection{Traditional image processing}
To evaluate the segmentation performance of the three vessel enhancement filters, we conducted experiments using two hyperparameter selection strategies. First, we performed a per-image grid search to identify the hyperparameter set that maximized the Dice similarity coefficient (DSC) for each image. This per-image optimization serves as an oracle upper bound, since it uses the ground-truth mask to select hyperparameters.

Second, we averaged the per-image optima across the dataset to obtain a single global configuration per filter. This mean-parameter setting removes the need for repeated tuning and provides a computationally efficient baseline. The resulting global parameters are summarized in Table~\ref{tab:results}.
\begin{table}[b]
    \centering
    \caption{Globally averaged hyperparameters for each vesselness pipeline, obtained by averaging the per-image oracle optima across the internal cohort. Reported values are means over per-image optima; discrete values are rounded to the nearest grid value during inference.}
    \begin{tabular}{|c|c|c|}
        \hline
        \textbf{Filter} & \textbf{Hyperparameter} & \textbf{Optimized Value} \\
        \hline
        \multirow{4}{*}{\textbf{Meijering}} 
            & \cellcolor[gray]{0.8}$\sigma$ & \cellcolor[gray]{0.8}3.76 \\\cline{2-3}
            & Threshold & 0.080 \\ \cline{2-3}
            & \cellcolor[gray]{0.8}Disk Size & \cellcolor[gray]{0.8}1 \\\cline{2-3}
            & Region Size & 2070 \\ \cline{2-3}
        \hline
        \multirow{4}{*}{\textbf{Sato}} 
            & \cellcolor[gray]{0.8}$\sigma$ & \cellcolor[gray]{0.8}4.14 \\\cline{2-3}
            & Threshold & 0.0195 \\ \cline{2-3}
            & \cellcolor[gray]{0.8}Mean Disk Size & \cellcolor[gray]{0.8}2.3 \\ \cline{2-3}
            & Region Size & 2128 \\ \cline{2-3}
        \hline
        \multirow{6}{*}{\textbf{Frangi}} 
            & \cellcolor[gray]{0.8}$\sigma$ & \cellcolor[gray]{0.8}2.5 \\\cline{2-3}
            & Threshold & 0.60 \\ \cline{2-3}
            & \cellcolor[gray]{0.8}$\alpha$ & \cellcolor[gray]{0.8}0.65 \\ \cline{2-3}
            & $\beta$ & 0.90 \\ \cline{2-3}
            & \cellcolor[gray]{0.8}Maximum Hole Area & \cellcolor[gray]{0.8}470 \\ \cline{2-3}
            & Minimum Region Size & 75 \\ \cline{2-3}
        \hline
    \end{tabular}
    \label{tab:results}
\end{table}

\begin{figure}[t]
  \centering
  \includegraphics[width=\textwidth]{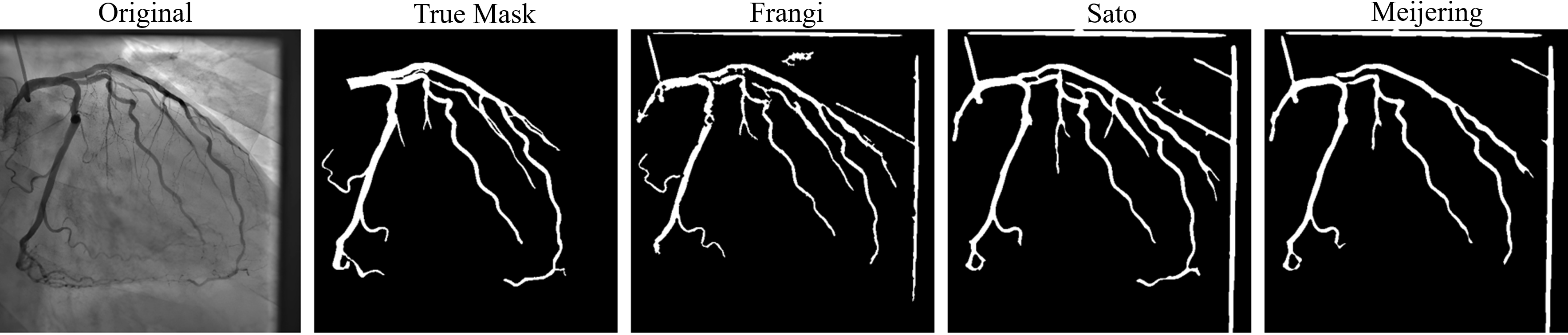}
  \caption{Representative qualitative outputs of classical vesselness-based pipelines on XCA frames. The examples illustrate differences in sensitivity to small branches and background artifacts, and the role of thresholding and morphological postprocessing in producing the final binary masks.}
  \label{fig:image_process}
\end{figure}

Table~\ref{tab:hyperparam_comparison} compares DSC under three strategies: oracle per-image optimization, a single global mean setting, and SVR-predicted per-image settings. An example output is shown in Figure~\ref{fig:image_process}.
\begin{table}[t]
    \centering
    \caption{Dice similarity coefficient (DSC) for classical vesselness pipelines under three hyperparameter-selection strategies: per-image oracle optimization (uses ground truth), a single global mean setting, and SVR-predicted per-image parameters (no ground truth at inference).}
    \label{tab:hyperparam_comparison}
    \resizebox{\textwidth}{!}{
    \begin{tabular}{|c|c|c|c|}
    \hline
    \textbf{Filter} & \textbf{Per-image Optimization (DSC)} & \textbf{Mean Parameters (DSC)} & \textbf{SVR-predicted Parameters (DSC)} \\
    \hline
    Meijering & 0.721 $\pm$ 0.027 & 0.698 $\pm$ 0.034 & \textbf{0.703$\pm$ 0.029} \\
    \hline
    Sato      & 0.735 $\pm$ 0.002 & 0.710 $\pm$ 0.031 & \textbf{0.719$\pm$ 0.027} \\
    \hline
    Frangi    & 0.783 $\pm$ 0.019 & 0.741 $\pm$ 0.024 & \textbf{0.759$\pm$ 0.018} \\
    \hline
    \end{tabular}}
\end{table}

\subsection{Improving image processing methods with machine learning}
Table~\ref{tab:hyperparam_comparison} shows that SVR-predicted hyperparameters consistently improve upon the global mean for all three filters. The absolute gains are +0.005 (Meijering), +0.009 (Sato), and +0.018 (Frangi), corresponding to relative improvements of approximately 0.7\%, 1.3\%, and 2.4\% over the mean-parameter baselines.

While the SVR approach remains below the oracle upper bound, it recovers a substantial portion of the mean-to-oracle gap: approximately 22\% (Meijering), 36\% (Sato), and 43\% (Frangi). The largest benefit is observed for Frangi, which is more sensitive to scale and threshold choices and therefore gains more from per-image adaptation.

These gains are achieved without access to ground-truth masks at inference. The SVR uses only inexpensive descriptors computed from the input image (intensity statistics, Minkowski profiles, and Betti curves), enabling per-image tuning while retaining the simplicity and efficiency of classical pipelines.

\subsection{Neural segmentation methods}
We next compare encoder--decoder CNNs (U-Net and FPN) and a Transformer baseline (Swin) under coronary-only supervision. Table~\ref{tab:internal_validation} reports internal-validation DSC (mean $\pm$ SD) for six model configurations.

\begin{table}[b]
  \centering
  \caption{Internal-validation Dice (mean $\pm$ SD) for neural segmentation architectures under coronary-only supervision, comparing backbone choice and input resolution.}
  \label{tab:internal_validation}
  \resizebox{\textwidth}{!}{
    \begin{tabular}{|c|c|c|c|c|}
      \hline
      \textbf{Segmentation Model} & \textbf{Backbone}    & \textbf{Input Size} & \textbf{\# Parameters (M} & \textbf{Internal-validation (DSC)} \\
      \hline
      U-Net             & SE-ResNet18        & 368$\times$368 & 20M & $0.865\pm 0.016$ \\
      \hline
      U-Net             & SE-ResNet18        & 756$\times$756 & 20M & $0.887\pm 0.013$ \\
      \hline
      U-Net             & SE-ResNeXt50       & 756$\times$756 & 35M & $0.905\pm 0.010$ \\
      \hline
      FPN (Best)        & SE-ResNet18        & 756$\times$756 & 16M & $\mathbf{0.914\pm 0.007}$ \\
      \hline
      FPN               & SE-ResNeXt50       & 756$\times$756 & 31M & $0.913\pm 0.008$ \\
      \hline
      Swin-Transformer  & -                  & 368$\times$368 & 30M & $0.889\pm 0.011$ \\
      \hline
    \end{tabular}
  }
\end{table}

Input resolution has a strong impact: U-Net improves from $0.865 \pm 0.016$ at $368\times368$ to $0.887 \pm 0.013$ at $756\times756$. Increasing encoder capacity yields smaller gains. At $756\times756$, upgrading U-Net from SE-ResNet18 to SE-ResNeXt50 improves DSC by about 0.018.

Architectural choice is more consequential in our setting. FPN at $756\times756$ attains the highest DSC, with FPN+SE-ResNet18 reaching $0.914 \pm 0.007$ and the SE-ResNeXt50 variant performing similarly ($0.913 \pm 0.008$). The smaller standard deviations for FPN suggest more stable performance across splits. Swin at $368\times368$ reaches $0.889 \pm 0.011$, outperforming U-Net at the same resolution and approaching U-Net at higher resolution, but it does not surpass the high-resolution FPN configurations.

\subsection{Effect of label design and external evaluation (DCA1)}
We evaluate the effect of supervision design using the best-performing architecture (FPN+SE-ResNet18 at $756\times756$). Training with merged coronary+catheter labels yields the highest internal-validation DSC: $0.931 \pm 0.006$ versus $0.914 \pm 0.007$ for coronary-only. Catheter-only supervision performs substantially worse ($0.856 \pm 0.031$), reflecting the difficulty of learning a robust device mask in isolation.

For external evaluation, we use the DCA1 cohort (Mexico). We consider two evaluation settings: (i) DCA1 used as an additional validation domain after light in-domain fine-tuning (``as external validation''), and (ii) DCA1 used as a strict external test without adaptation (``as external test''). Table~\ref{tab:validation_by_labeling} summarizes results.

\begin{figure}[t]
  \centering
  \includegraphics[width=0.6\textwidth]{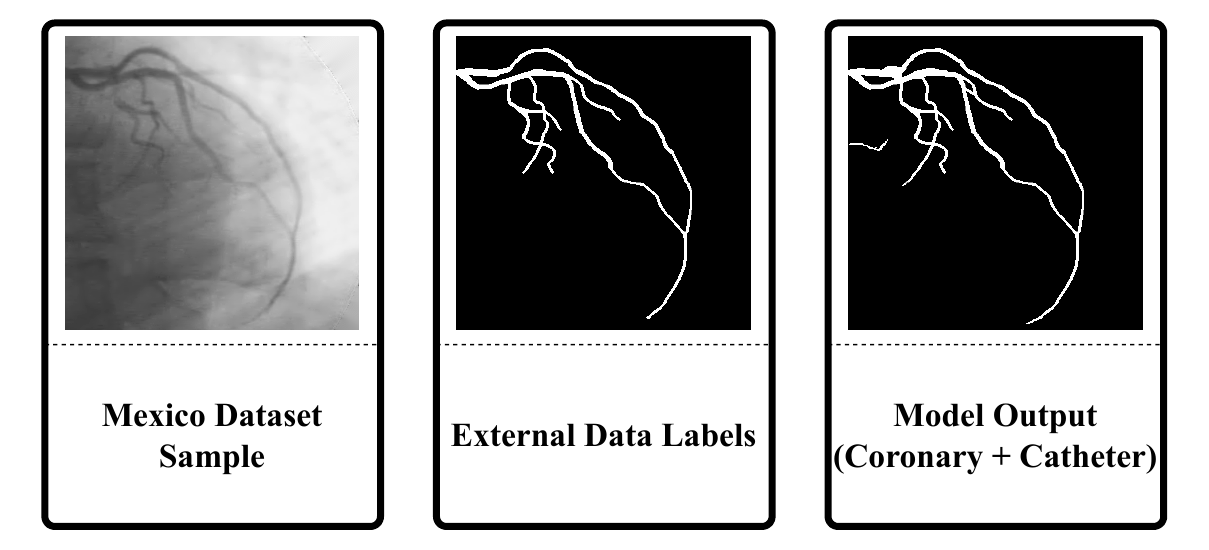}
  \caption{Example prediction on the external DCA1 cohort compared with the provided ground-truth vessel mask. This figure highlights typical cross-dataset differences in appearance and labeling conventions that contribute to performance drops under strict external testing.}
  \label{fig:3_Mexico_dataset}
\end{figure}

As a strict external test, performance drops relative to internal validation, indicating domain shift. Coronary-only falls from 0.914 to $0.798 \pm 0.029$, and merged supervision falls from 0.931 to $0.814 \pm 0.026$. With light in-domain fine-tuning, both settings recover to approximately $0.881 \pm 0.014$ (coronary-only) and $0.882 \pm 0.015$ (merged). A qualitative example on the external DCA1 cohort, illustrating typical cross-dataset appearance and labeling differences, is shown in Figure~\ref{fig:3_Mexico_dataset}.

\begin{table}[ht]
  \centering
  \caption{Dice on internal validation and on the external DCA1 cohort under different label regimes (coronary-only, catheter-only, and merged coronary+catheter). Results are reported for DCA1 used as an external test and after light in-domain fine-tuning.}
  \label{tab:validation_by_labeling}
  \resizebox{\textwidth}{!}{
    \begin{tabular}{|c|c|c|c|}
      \hline
      \textbf{Labeling Method} & \textbf{Internal-validation} & \textbf{Mexico (with adaptation)} & \textbf{Mexico (no adaptation}) \\
      \hline
      Coronary          & $0.914 \pm 0.007$ & $0.881 \pm 0.014$ & $0.798 \pm 0.029$ \\
      \hline
      Catheter          & $0.856 \pm 0.031$ & --              & --              \\
      \hline
      Coronary+Catheter & $0.931 \pm 0.006$ & $0.882 \pm 0.015$ & $0.814 \pm 0.026$ \\
      \hline
    \end{tabular}}
\end{table}

\subsection{Vessel-type classification (LAD/LCX/RCA)}
\begin{figure}[!b]
  \centering
  \includegraphics[width=0.9\textwidth]{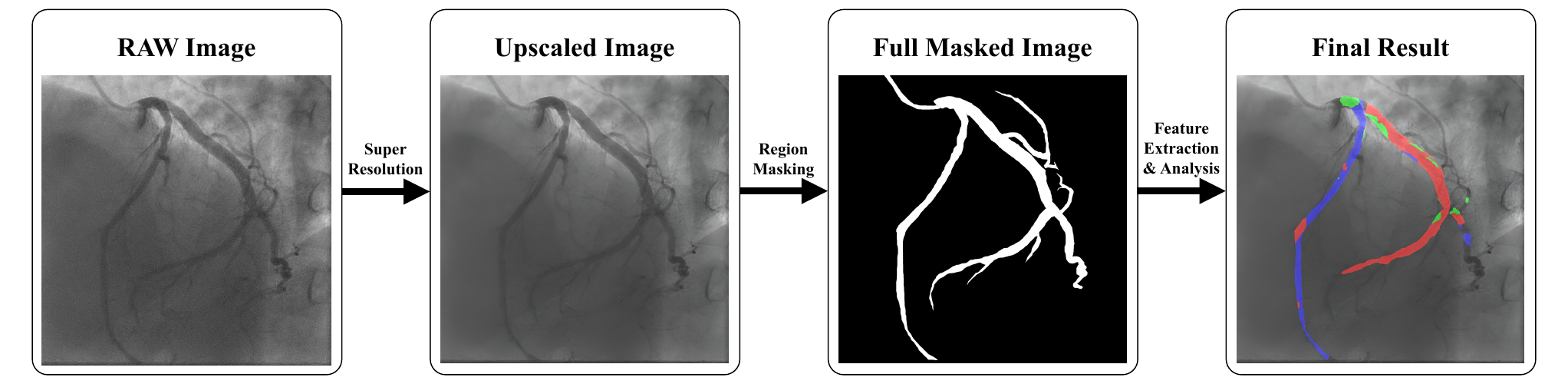}
  \caption{Qualitative vessel-type labeling within the segmented coronary tree. Pixels in the first-stage foreground are assigned to one of the main coronary classes by the second-stage model; colors indicate the predicted vessel identity.}
  \label{fig:4_classes}
\end{figure}

Building on the best first-stage segmenter, we trained a shared backbone with three vessel-specific heads to assign vessel identity within the predicted foreground. A qualitative example of the vessel-type labeling outputs is shown in Figure~\ref{fig:4_classes}. At inference, per-pixel posteriors for LAD, LCX, and RCA are computed and gated by the first-stage mask. Each foreground pixel is assigned the argmax class, followed by small-component pruning and contour smoothing. Representative qualitative results of the LAD/LCX/RCA assignment are presented in Figure~\ref{fig:4_classes}.

Table~\ref{tab:accuracy_by_coronary_type} summarizes performance by vessel type. RCA attains the highest scores, consistent with its more isolated course in common projections. Most residual errors for LAD and LCX concentrate near the left main bifurcation and in overlapping segments, where foreshortening and contrast heterogeneity complicate separation.
\begin{table}[!b]
  \centering
  \caption{Vessel-type labeling performance within the coronary tree, reported as pixel accuracy and Dice (DSC) for LAD, LCX, and RCA.}
  \label{tab:accuracy_by_coronary_type}
  \begin{tabular}{|c|c|c|}
    \hline
    \textbf{Coronary Type} & \textbf{Accuracy (\%)} & \textbf{DSC} \\
    \hline
    LAD & $95.4 \pm 1.6$ & $0.786 \pm 0.031$\\
    \hline
    LCX & $96.2 \pm 1.1$ & $0.794 \pm 0.030$\\
    \hline
    RCA & $98.5 \pm 0.6$ & $0.844 \pm 0.022$\\
    \hline
  \end{tabular}
\end{table}

\section{Discussion and Conclusion}
\label{sec:dis}

\paragraph{Overall summary of contributions.}
This work presents an end-to-end pipeline for coronary analysis in X-ray coronary angiography (XCA), spanning dataset curation, segmentation, and vessel-type labeling. We provide an expert-reviewed cohort with both global masks (coronary arteries and catheter) and vessel-level annotations for the major coronaries (LAD, LCX, RCA). This enables evaluation beyond binary vessel segmentation and supports vessel-specific downstream analyses.

We benchmark classical vesselness pipelines (Meijering, Frangi, Sato) against modern deep models under a consistent protocol. For classical methods, we introduce a learning-to-tune strategy that predicts filter hyperparameters on a per-image basis from inexpensive descriptors, improving performance without requiring ground-truth labels at inference time. For deep segmentation, we compare U-Net, FPN, and a Transformer baseline, and we analyze the effect of supervision design by training with coronary-only, catheter-only, and merged coronary+catheter labels. Finally, we evaluate external transfer on DCA1 to characterize domain shift and the impact of modest in-domain adaptation.

\paragraph{Classical image processing under limited labels.}
Classical vesselness pipelines remain useful when labeled data are scarce. They incorporate strong inductive bias for tubularity via Hessian eigen-structure, require minimal training, and provide interpretability and efficiency. In our experiments, learning to predict filter hyperparameters from compact image descriptors recovers a meaningful portion of the oracle performance advantage without relying on labels at inference. The strongest learned classical pipeline, Frangi with SVR-predicted parameters, achieves a DSC of 0.759, compared with the per-image oracle bound of 0.783 (Table~\ref{tab:hyperparam_comparison}).

These results do not aim to replace modern deep segmenters on sufficiently large labeled cohorts. Instead, they support classical pipelines as strong, data-efficient baselines and practical alternatives when data, compute, or deployment constraints make end-to-end training difficult.

\paragraph{Neural segmentation: accuracy and practical trade-offs.}
Among deep models, both multi-scale fusion and high spatial resolution were important for performance. FPN with an SE-ResNet18 encoder at $756\times756$ achieved $0.914 \pm 0.007$ under coronary-only supervision and $0.931 \pm 0.006$ under merged coronary+catheter supervision (Tables~\ref{tab:internal_validation} and \ref{tab:validation_by_labeling}). These results substantially exceed the classical baselines in our setting.

The Swin-Transformer baseline at lower resolution performed competitively with higher-resolution U-Net, suggesting that global context helps in angiographic segmentation. However, the highest accuracy and stability were achieved by the high-resolution FPN configurations. In practice, FPN+SE-ResNet18 provides a favorable accuracy--efficiency balance, while heavier backbones yield smaller incremental gains.

\begin{figure}[t]
  \centering
  \includegraphics[width=0.78\textwidth]{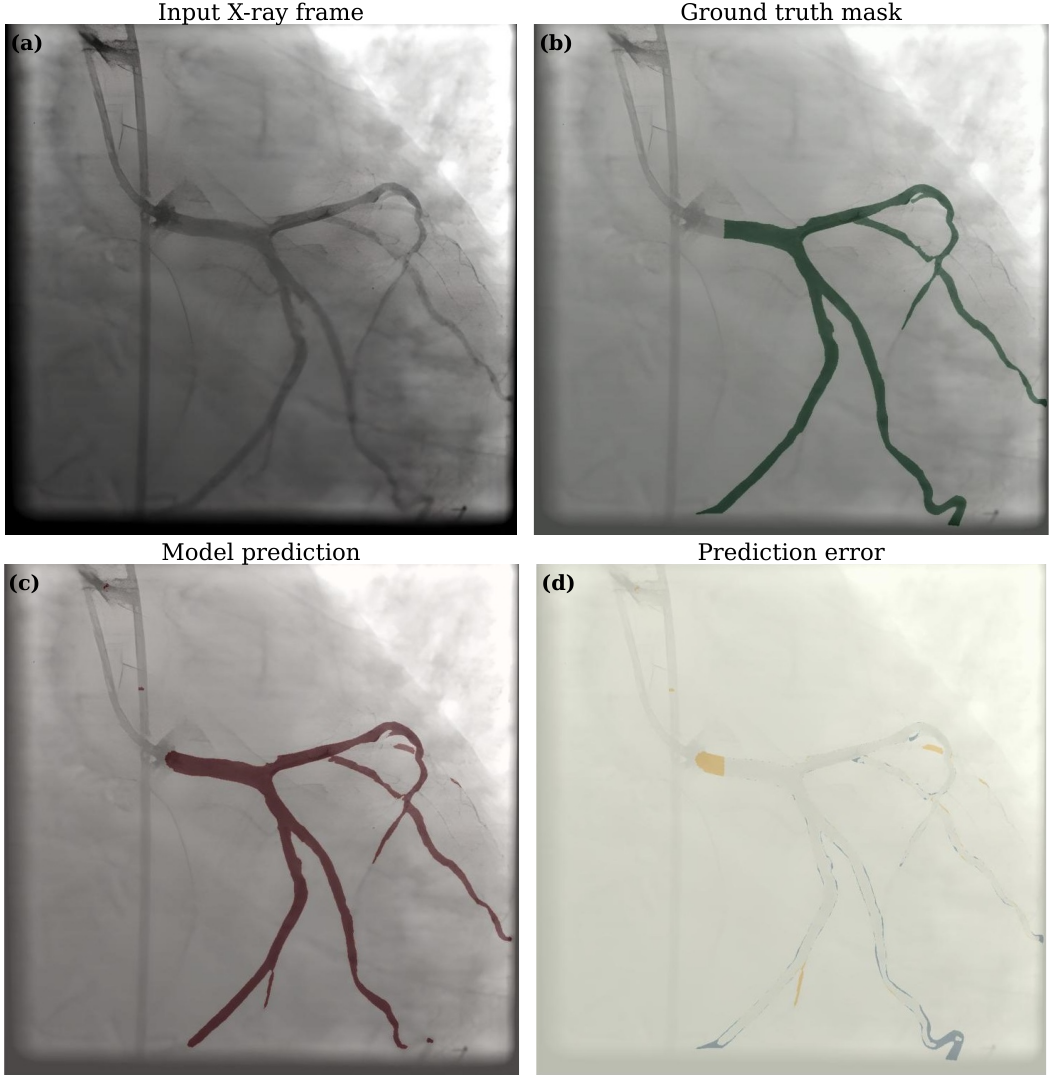}
  \caption{Error analysis of first-stage segmentation. Shown are an input frame, the reference mask, the model prediction, and an error map that separates false positives and false negatives. The error map supports qualitative inspection of common failure modes near low-contrast segments, overlaps, and catheter-adjacent regions.}
  \label{fig:error}
\end{figure}

\paragraph{Error analysis and the role of label design.}
Residual segmentation errors were most common near the left main bifurcation and in overlapping segments such as diagonals and obtuse marginals, where foreshortening and heterogeneous contrast reduce separability. This pattern is reflected in vessel-type performance, where RCA achieved higher DSC than LAD and LCX (Table~\ref{tab:accuracy_by_coronary_type}). Errors also occur around catheter-adjacent regions, where boundary ambiguity can lead to over-segmentation or missed vessel pixels. An error map separating false positives and false negatives for a representative case is presented in Figure~\ref{fig:error}.

Merged coronary+catheter supervision improved robustness and reduced variability on internal validation. One plausible explanation is that treating catheter as foreground during training reduces ambiguity at coronary--device boundaries, allowing the model to learn a more consistent notion of foreground structure. When evaluating on DCA1, domain shift leads to an expected drop in DSC under both coronary-only and merged settings. Light in-domain fine-tuning recovers a large portion of this gap (Table~\ref{tab:validation_by_labeling}), suggesting that modest adaptation can be sufficient to align to new acquisition and labeling standards.

\paragraph{Vessel-type labeling and relation to downstream analysis.}
Vessel-type classification of LAD, LCX, and RCA enables vessel-specific morphometric analysis, including reference diameter, percent diameter stenosis, and lesion length, as well as territory-specific downstream modeling. These measurements are relevant for developing non-invasive surrogates of functional assessment, including FFR-related analyses, when combined with appropriate centerline extraction and hemodynamic modeling assumptions.

In our experiments, the vessel-type classifier achieved high accuracy and DSC across all three classes, with the strongest performance on RCA (Table~\ref{tab:accuracy_by_coronary_type}). Most residual confusions occurred near the left main bifurcation and in regions of overlap, where delineating LAD versus LCX is inherently challenging from a single projection.

\paragraph{Limitations and practical considerations.}
This study has several limitations. First, ground-truth annotations do not exhaustively cover all side branches. Low-contrast and thin vessels remain challenging for both classical and neural methods. Second, our internal cohort is collected from a single center under a specific acquisition protocol. Although the DCA1 experiments indicate some transferability, broader multi-center evaluation is needed to assess generalization across detectors, frame rates, and contrast protocols.

Third, we select a single best frame per cine and do not use temporal information. Leveraging short sequences could help address motion, contrast propagation, and overlap. Fourth, we do not explicitly impose vascular-tree topology constraints, which may reduce spurious connections and improve structural consistency. Finally, while merged-label training improves robustness, some downstream applications may require explicit device suppression. In such cases, a dual-head model that produces separate coronary and device masks may be preferable.

\paragraph{Future work.}
Several directions merit further exploration. Incorporating short cine sequences using 2.5D models or temporal transformers may better exploit contrast dynamics and motion cues. Semi-supervised and self-supervised pretraining on unlabeled cines may reduce annotation requirements and improve feature robustness. Domain generalization and test-time adaptation strategies may further improve cross-center performance.

In addition, topology-aware decoders or graph-regularized losses could enforce vascular-tree structure and reduce spurious artifacts. Active learning workflows could identify high-uncertainty regions for targeted expert labeling. Finally, integrating vessel-type labels with efficient hemodynamic surrogates may help connect anatomical segmentation with functional assessment in routine workflows.

\paragraph{Concluding remarks.}
In summary, two complementary strategies, learning-enhanced classical image processing and deep neural segmentation, support accurate coronary segmentation and vessel-type labeling in XCA. Learned per-image tuning strengthens classical baselines in low-label settings, while high-resolution multi-scale models provide strong performance on our internal cohort and recover with modest adaptation on external data. The resulting masks and vessel labels provide a practical foundation for vessel-specific coronary analytics and downstream research applications.

\label{sec:acknowledgement}
\section*{Acknowledgement}
The authors extend their gratitude to the physicians and clinical staff at the Shahid Rajaie Cardiovascular Research and Treatment Institute, Tehran, Iran, for their support, and Dr. Farideh Beyki for helpful discussions. Ethical approval for this study was obtained from the same institution (protocol code IR.RHC.REC.1403.038).

The authors also acknowledge the use of \emph{ChatGPT} (\emph{OpenAI}, \href{https://chatgpt.com/}{chatgpt.com}) to assist with language editing and manuscript readability. All research ideas, analysis, and conclusions are solely the responsibility of the authors.

\bibliographystyle{unsrtnat}
\bibliography{references}  

\end{document}